\documentclass{svproc}
\usepackage{url}
\usepackage{comment}
\usepackage[pagebackref,breaklinks,colorlinks,citecolor=cvprblue]{hyperref}

\usepackage{graphicx}
\usepackage{amsmath}
\usepackage{amssymb}
\usepackage{booktabs}

\usepackage[ruled, lined, longend]{algorithm2e}
\usepackage{colortbl}
\usepackage{xcolor}
\usepackage{color}
\usepackage{amsmath}
\usepackage{lipsum}

\usepackage{wrapfig}

\usepackage{censor}
\usepackage{tikz}
\usepackage{pgfplots}
\usepackage{svg}
\usetikzlibrary{pgfplots.groupplots}
\usepgfplotslibrary{fillbetween}
\usepackage{adjustbox}
\usepackage{indentfirst}
\usepackage{tabularx}
\usepackage{bm}

\usepackage{comment}
\usepackage{censor}
\usepackage{tikz}
\usepackage{pgfplots}
\usepackage{svg}
\usetikzlibrary{pgfplots.groupplots}
\usepgfplotslibrary{fillbetween}
\usepackage{adjustbox}
\usepackage{indentfirst}
\usepackage{tabularx}
\usepackage{subcaption}

\usepgfplotslibrary{groupplots}

\usepackage{float}

\definecolor{cvprblue}{rgb}{0.21,0.49,0.74}

\begin{document}
\mainmatter              
%
\title{Efficiently Linking Real Scenes with Synthetic Data Generation for AI-based Cognitive Robotics and Computer Vision Applications}

\titlerunning{Efficiently Linking Real Scenes with Synthetic Data Generation}  
%
\author{Paul Koch\textsuperscript{1},  Vivek Chavan\textsuperscript{1}, André Sers\textsuperscript{1}, Adem Karakurt\textsuperscript{1}, Paul Hofmann\textsuperscript{1}, Mohamad Zaher Ziadeh\textsuperscript{1}, and Jörg Krüger\textsuperscript{1,2} \\
{\tt\small \
 Website: \href{https://www.ipk.fraunhofer.de/}{www.ipk.fraunhofer.de}}}
\authorrunning{Koch et al.} 
%
%
\institute{Fraunhofer IPK, Berlin, Germany.\\
\and
TU Berlin, Germany}

\maketitle              

\begin{abstract}
AI vision models are a driving factor for the potential use case scenarios of cognitive robotics within in the industry and household applications. A large array of methods from semantic environment analysis towards 6D and grasping pose estimation have been proposed based on the latest AI achievements. 
However, such advancements require further strong and efficient methods w.r.t. training data and AI-architectures, which are capable in synergy to tackle current challenges, precision limits, and scalability beyond domain gaps. In this paper, we discuss these current limits and trends in the related state-of-the-art which are challenging those. Further we discuss our current work in progress on bridging the domain gap between simulations and real world applications by linking those in the training data generation. 

\end{abstract}

\section{Introduction}
In recent years, AI-methods have shown significant advancements in various data-driven applications, surpassing traditional computer science methods. Initially successful in computer vision in 2012~\cite{alexnet}, AI has now made significant progress in robotic-related vision applications. Many methods have been proposed to detect objects~\cite{FasterRCNN,Yolo1,DETR,DeformDETR,CoDetr},
including pixel-wise panoptic segmentation of the environment and objects~\cite{FasterRCNN,DETR,Mask2Former,CoDetr}. 
Additionally, AI has made advancements in 6D pose estimation and grasping pose estimation of partially occluded objects~\cite{poseCNN-YCBV,maskFusion,DenseFusion,ZebraPose,AnyGrasp}
. Overall, AI has demonstrated its potential to generalize with self-supervised learning towards various data-driven downstream applications with optional task specific fine-tuning~\cite{DinoV2,clip,GPT3}.\\
\indent However, the success of AI-methods in robotic-related applications still heavily relies on the availability of well-annotated large datasets with high diversity to ensure generalizability within a certain task and domain. To address this need, numerous datasets and challenges for robotic-related applications have been introduced. Among these, notable pioneers for end2end AI-models in their respective challenges are YCB-Video~\cite{poseCNN-YCBV} and Linemod~\cite{LineMod} for 6D pose estimation, and GraspNet~\cite{GraspNet1B} for two-finger grasping pose estimation. Other robotic manipulation centered challenges such as SuchtionNet~\cite{suctionnet}, 6D pose estimation of unseen objects~\cite{unseen6D}, and Grasping transparent objects~\cite{TransparentCompGrasp} have been proposed. Additionally, some work further focuses on incorporating simulations~\cite{poseCNN-YCBV,SSL6D,DataGenSimDexNet2,DataGenSimGraspNet,DataGenSimJacquard,DataGenSimGeometry-Aware,kochSyntDataMMM}
and robots~\cite{DataGenRobotSSl,DataGenRobotHandEye,SSL6D,Koch6DGen} for unlimited synthetic data generation to tackle the high demands of precise annotated training data. Though they come with issues of the domain gap~\cite{domain_gap_1995}, execution speed~\cite{GraspNet1B}, and implementation costs.\\
\indent W.r.t robot control policy learning various methods have been proposed. Driess et al.~\cite{PaLME} created with PalmE a multi modal model for language controlled robot action, decision and motion planning, which is capable to grasp objects on request. This innovation is based on heavy parallel robotic training data collection, where hard coded robot actions are observed based on a natural language text requests. Similarly, Chi et al.~\cite{DiffusionPolicy} observe use human controlled robot actions in order to learn end2end manipulation policies. W.r.t reinforcement learning, further control policy learning are proposed for simulated data~\cite{metaWorld,GradientSurgery,RLContext-based,HumandInTheLoop}. Anyhow, these methods, datasets and challenges play a crucial role in advancing the capabilities of AI in robotic vision applications by providing standardized evaluation benchmarks and fostering research innovations.\\
\indent Ever since various AI-architectures~\cite{ZebraPose,maskFusion,SSL6D,GraspNess,AnyGrasp,DynamicGrasp,DenseFusion,DexNet4-2019}
with hard-coded engineering methods (tailored methods for their specific problems) have been proposed, which are further increasing precision on their related downstream tasks and benchmarks. One can increasing consider in particular Bin-Picking challenges to be solved~\cite{AnyGrasp,DexNet4-2019}. Here AnyGrasp~\cite{AnyGrasp} shows a high resilience on bin-picking various objects which can be scaled towards unknown and dynamic objects. Although AnyGrasp is versatile, it is still limited to a semi-2D top-down grasping setup with a two finger gripper or suction end-effector~\cite{suctionnet,DexNet4-2019} and not proven in many industrial setting. Moreover, these current grasping specific methods lack generally knowledge about the environment and the objects within the scene, making them powerful modular tools but not scalable towards full AI end-to-end reasoning-based manipulation planning - rather they are a streamlined end-to-end mapping of the statistical source domain knowledge to the target domain of; grasping poses~\cite{GraspNet1B,DexNet4-2019,DynamicGrasp}, grasping rectangle areas~\cite{RecanlgeGrasp2011,RecanlgeGrasp2015,DataGenSimDexNet2,RecanlgeGrasp2018}, grasping affordance maps~\cite{GraspNess,GraspAffordance,Affordance2020}, 6D pose estimation attached with predefined grasping poses~\cite{poseCNN-YCBV,DenseFusion,Koch6DGen,SSL6D}, or robot control policies~\cite{PaLME,DiffusionPolicy}. One may argue that those robot control policies~\cite{PaLME,DiffusionPolicy} are reasoning-based methods, but they are still limited towards their training domain and lack explicit physical world knowledge such as "what is object mass", "whats a center of mass", "whats stiffness", "whats friction", etc. Following this logic, Huang et al.~\cite{DefGraspSim} improved grasping of deformable objects based on a physic-reasoning enhanced backbone.\\
\indent In this paper we continue this logic and argue that achieving a generalized and physics grounded reasoning capable system would require a more "cognitive" AI robotic model rather then many individually engineering tailored service modules - and in order to achieve such a AI-system extensive multi modal well annotated data is required. Building upon pioneers related work, we propose a work-in-progress data collection and AI-training loop, which links real world scenes with simulations for unlimited and well annotated real data.   

\section{Related Work}
\indent \textbf{Hand held data generation:} The most strait forward method for vision-based cognitive robotic centered data generation is to use single images and videos for data collection. This works fine for classical vision problems such as classification~\cite{imagenet}, detection~\cite{msCOCO} and segmentation~\cite{cityScapes,msCOCO}. However, annotations such as poses for cognitive robotic tasks are much harder to annotate. Therefore, related work often use video streams with camera trajectory tracking, such that poses only need to be annotated once per scene~\cite{poseCNN-YCBV,LineMod,HomebrewedDB,BOP}. This creates a dense data collection but is limited to a few scenes.\\
\indent \textbf{Data generation with robots:} In addition to hand held data collection, other work uses robotic arms in order to roam the work space and track the camera positions~\cite{GraspNet1B,SSL6D,Koch6DGen}. This method gives a marker free scene with precise camera trajectory tracking within the target robot workspace. However, this setup is also limited to a few scenes. Therefore, Deng et al.~\cite{SSL6D} proposed a data collection loop within which the robot can iteratively learn to move objects in the scenery. Hence, generation many constellations to refine the performance, but with a limited object variety. Furthermore, these methods are further time restricted and scalable only with a large invest. In contrast to pose estimation datasets, other work proposes to observe the hard-coded or human controlled robot actions in order to directly learn robot control and manipulation policies~\cite{PaLME,DiffusionPolicy}. However, these methods are also bound by extensive investment costs and their training domain.\\
\indent \textbf{Data generation with simulations:} In order to create well annotated training data with scene and object diversity, related work opted to employ simulations for unlimited synthetic data generation~\cite{LineMod,poseCNN-YCBV,HomebrewedDB,SSL6D}. Other work uses simulators as a source for annotating grasping poses~\cite{GraspNet1B}. However, these methods suffer from the domain gap~\cite{domain_gap_1995} when shifted to the target domain~\cite{SSL6D}. Although methods such as domain randomization~\cite{Domain_Randomization_2017} and style transfer~\cite{StyleTransfer,DomainEnhancedStyleTransfer} are trying to bridge the domain gap, synthetic data remains to be a supporting data source along real data~\cite{SSL6D,kochSyntDataMMM}.\\
\indent \textbf{Grasping Annotations:} Unlike other computer-vision annotations such as classification labels, bounding boxes, segmentation masks, and none symmetric 6D object poses are grasping poses not of discrete nature. The realm of possible grasping poses are unlimited in the continuous cartesian space~\cite{DataGenRobotSSl}. Hence, learning regression methods for estimation is an ill defined problem. Nevertheless, related work pursed this avenue with success~\cite{GraspNet1B}. In order to deal with this discrete vs. continuous problem some work focuses on introducing grasping affordance~\cite{GraspAffordance,Affordance2020}, none-discrete matrices~\cite{poseCNN-YCBV}, or offset tolerance based loss functions~\cite{GraspNet1B}. Learning from demonstration~\cite{PaLME,DiffusionPolicy} and reinforcement learning~\cite{metaWorld,GradientSurgery,RLContext-based,HumandInTheLoop} is circumventing this issue by directly learning the robot control policy. But here learning from demonstration is limited to the discrete number of demonstrations, while reinforcement learning is mostly bound to simulations and has a hard time to converge for complex problems without constrains or human feedback~\cite{HumandInTheLoop}.

\section{From Real Scenes to Simulations and Back}
\begin{figure*}
    \centering
    \includegraphics[width=\textwidth]{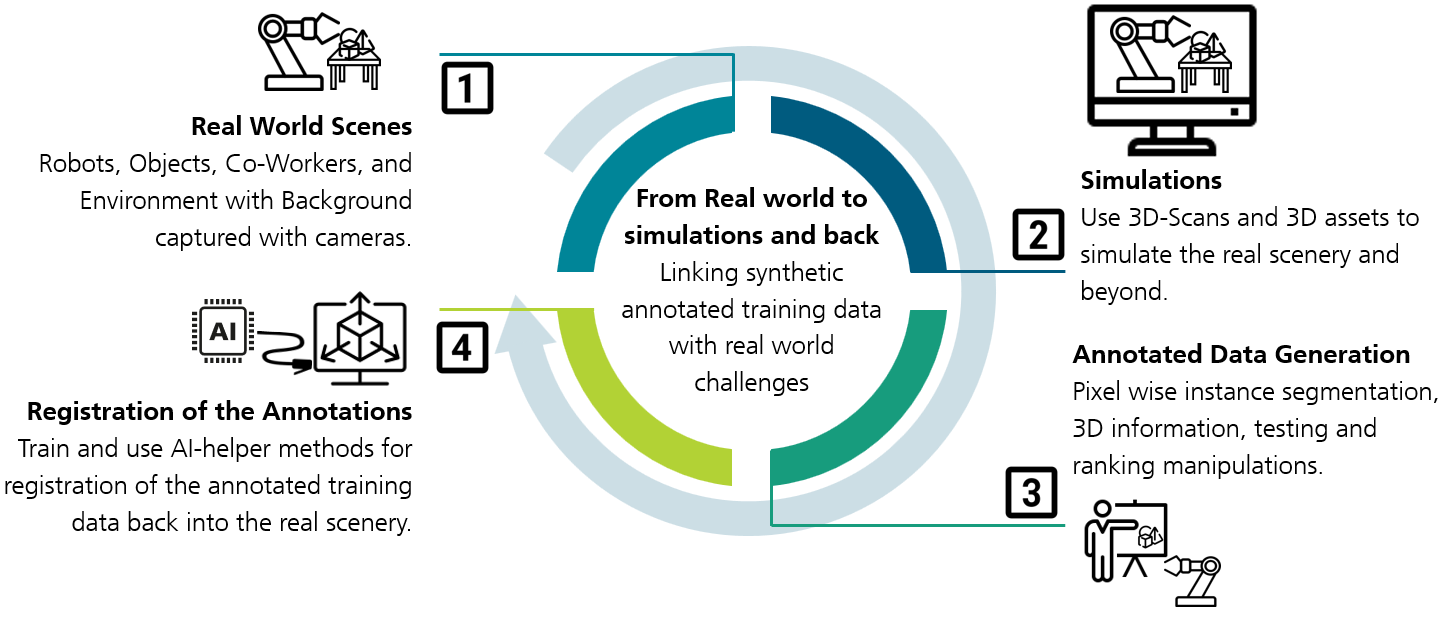}
    \caption{\textbf{Linking real robot workspace scenes with simulations:} In a continuous loop we propose to scan real scenes (1) and transform them into simulations (2). Here we can conduct many experiments, find grasping candidates, train control policies and annotate training data (3). Eventually, we can now train further AI methods to help to transform the annotations and control policies back into the real scenery (4). This process is looped in order to refine AI-helpers and deploy methods.}
    \label{fig:pipeline}
\end{figure*}

We outline our main data generation pipeline in Fig.~\ref{fig:pipeline}. 

\begin{figure}
    \centering
\begin{minipage}{1.0\textwidth}
  \centering
\subcaption{Three view examples.}
\includegraphics[width=1.0\textwidth]{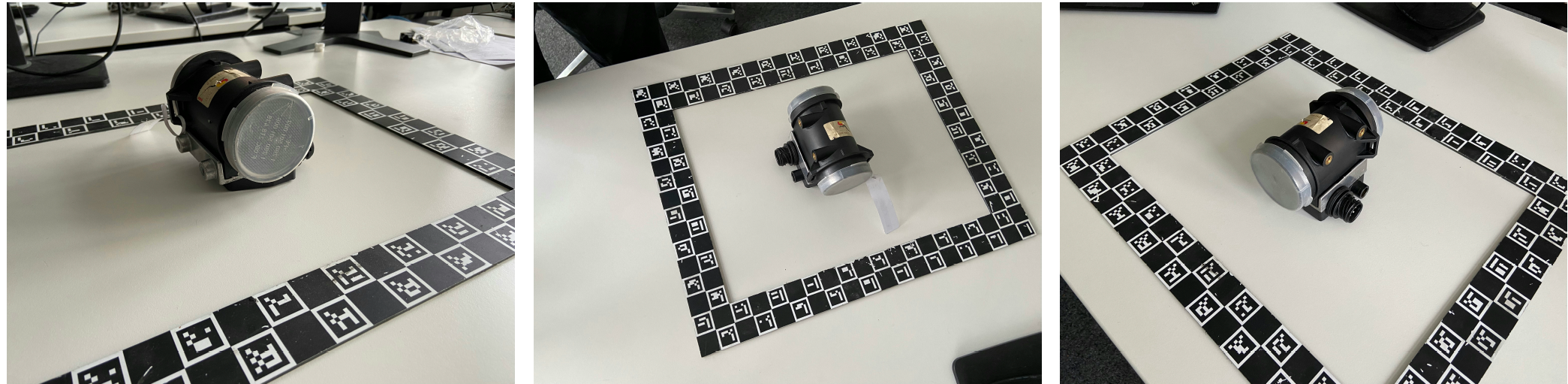}
\label{fig:1a}
\end{minipage}
\begin{minipage}[b]{0.25\textwidth}
  \centering
\includegraphics[width=1.0\textwidth]{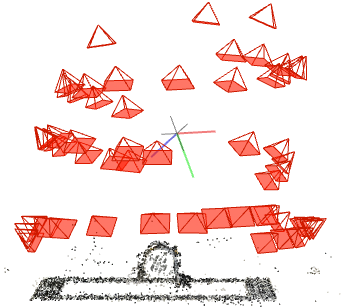}
\subcaption{All Views}\label{fig:1b}
\end{minipage}
\begin{minipage}[b]{0.3\textwidth}
  \centering
\includegraphics[width=1.0\textwidth]{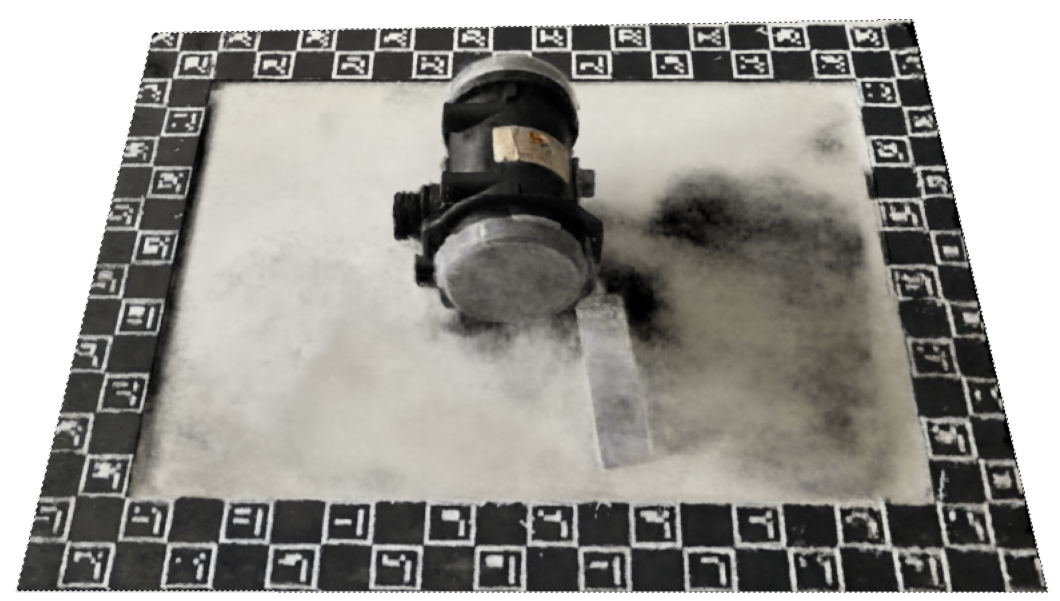}
\subcaption{Nerf}\label{fig:1c}
\end{minipage}
\begin{minipage}[b]{0.25\textwidth}
  \centering
\includegraphics[width=1.0\textwidth]{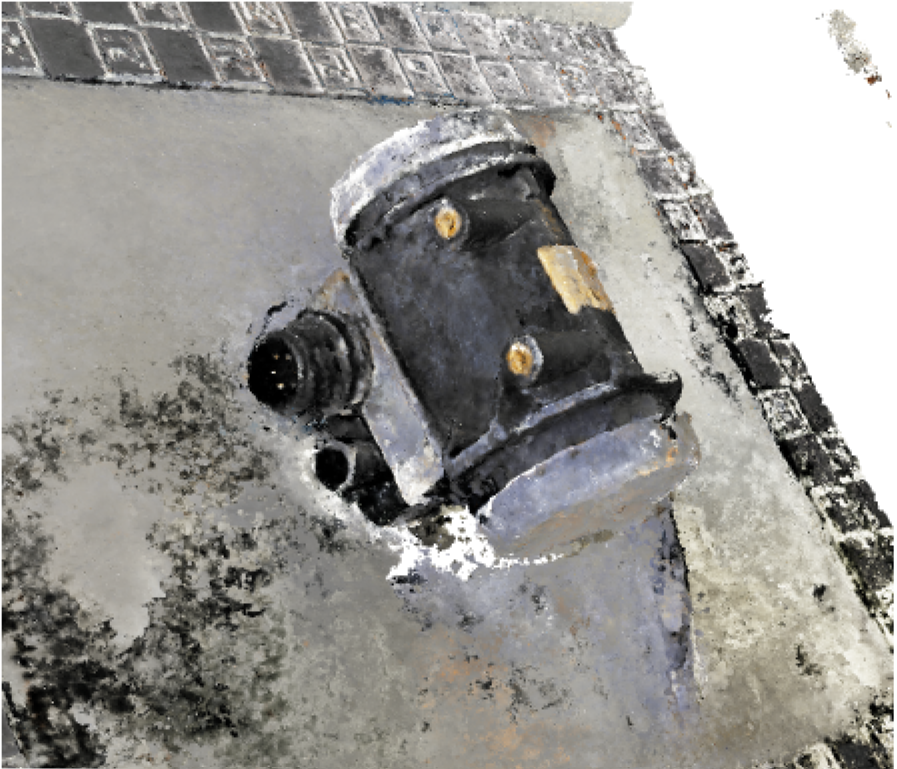}
\subcaption{CAD}\label{fig:1d}
\end{minipage}
\caption{With Nerfs~\cite{InstantNGP} we can create good looking 3D representations (some issues with the shade), but we cant yet export high quality masked 3D metric assets and textures.}
\label{fig:nerf}
\end{figure}

\begin{figure}[h]
    \centering
\begin{minipage}{0.17\textwidth}
  \centering
\includegraphics[width=1.0\textwidth]{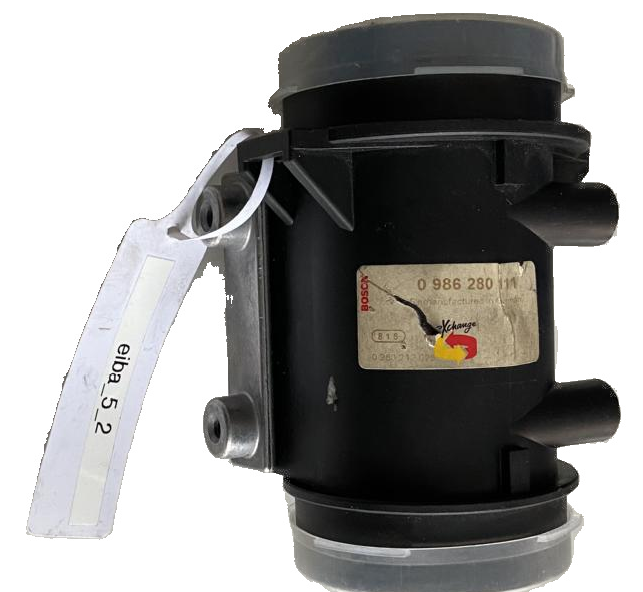}
\subcaption{Image}\label{fig:2a}
\end{minipage}
\begin{minipage}{0.17\textwidth}
  \centering
\includegraphics[width=1.0\textwidth]{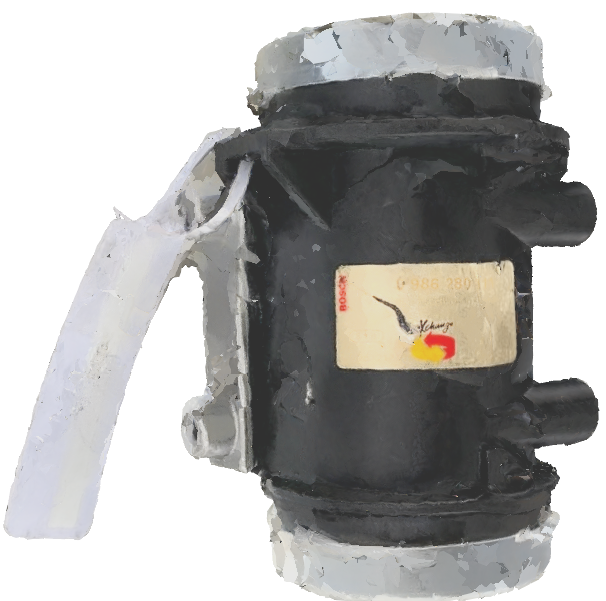}
\subcaption{Mesh}\label{fig:2b}
\end{minipage}
\begin{minipage}{0.17\textwidth}
  \centering
\includegraphics[width=1.0\textwidth]{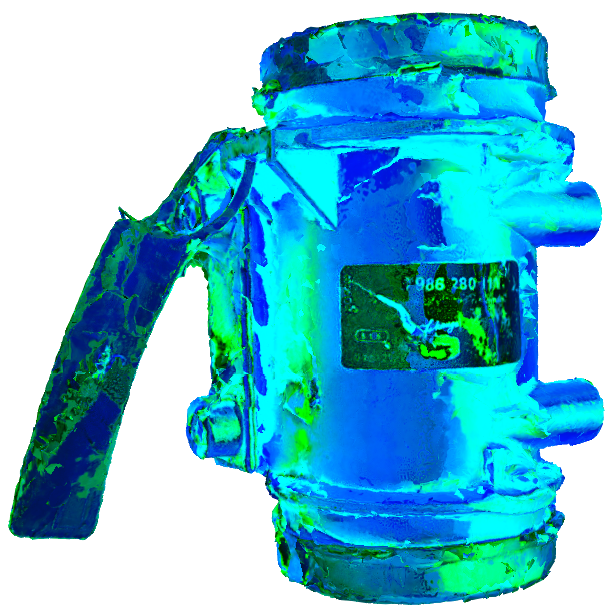}
\subcaption{Surface}\label{fig:2c}
\end{minipage}
\begin{minipage}{0.17\textwidth}
  \centering
\includegraphics[width=1.0\textwidth]{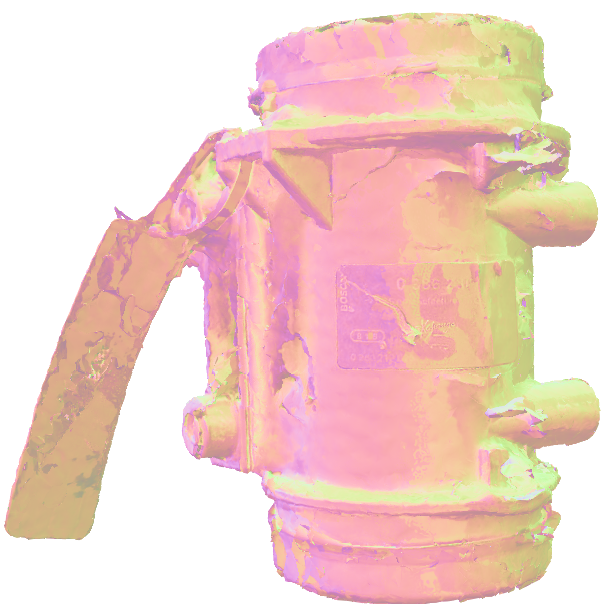}
\subcaption{Normals}\label{fig:2d}
\end{minipage}
\begin{minipage}{0.17\textwidth}
  \centering
\includegraphics[width=1.0\textwidth]{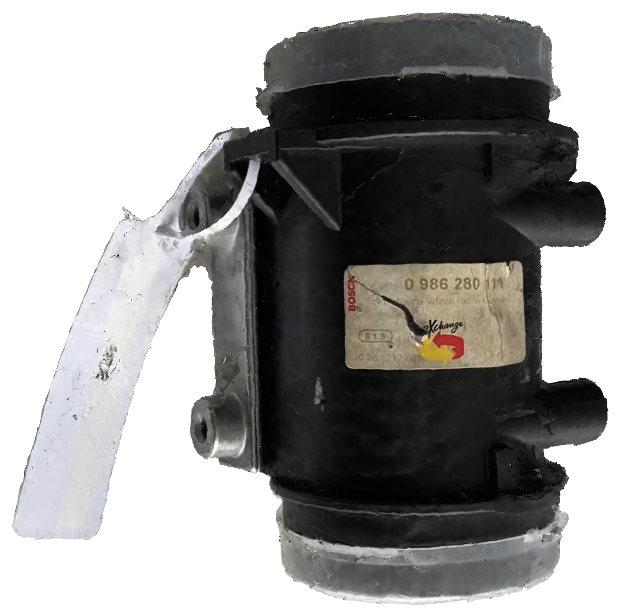}
\subcaption{Render}\label{fig:2e}
\end{minipage}
\caption{3D Assets with Textures from Nvdiffrec~\cite{nvdiffrec1}}
\label{fig:cad}
\end{figure}

\textbf{(1) Scanning the Scenery:} Creating simulations requires many assets and textures in order to represent the real world scenery. Therefore, scanning the scenery assets has been proposed by the research community rather then using time consuming and expert crafting required CAD modelling. Recent advancements in AI research further improved scanning methods, such that no expensive hardware is required. Here Nerfs have been found to be well suited to reconstruct 3D scenes~\cite{InstantNGP} (see Fig.~\ref{fig:nerf}), but it is hard to convert them into mesh and texture assets of individual objects for simulations. In contrast other scanning methods such as Nvdiffrec~\cite{nvdiffrec1,nvdiffrecmc2} allow us to directly get those assets disentangled from the background~\ref{fig:cad}. For the scenery itself nvdiffrec can also be used, but this is not yet thoroughly investigated. 

\textbf{(2) Generate Simulations:} From the 3D assets we can now build novel simulated scenarios with randomly placed objects. However the scene generation needs to incorporate prior knowledge about the possible constellations of things, such that e.g. the objects are reasonably placed on tables and in shelves rather then on the floor. While it is fairly easy to hardcode this on a small scale, it becomes quite a challenge scaling to generate well structured simulations without human controlled design in order to cut costs. Here one needs assets which are connected to meta information on how a simulation can use this assets within a full autonomous scene generation -- a table asset has object placement areas and object assets require such area and are graspable. A full logic needs to be developed here.  


\textbf{(3) Annotating Data:}
Modern simulators such as Omniverse\texttrademark{} and Blender are capable to fully randomize their assets and create photorealistic images from their generated scenes. Moreover, these simulators are capable to extract segmentation annotations and other meta information such as point clouds and 6D poses from their scenes. In addition to logical placements of assets in the scene, a simulation also need to incorporate reasonable lighting and physics (e.g. gravity). Nvidia Issac Sim\texttrademark{} is a recent development within simulators, which allows users to create and generate robotic scenes. Building on top of Omniverse\texttrademark{} this simulator delivers photorealistic and physically accurate virtual environments. Additionally, the Issac Sim\texttrademark{} simulator can be used to carry out many robotic experiments. Hence, one can use it in order to generate for a given scene e.g. possible grasping poses with force requirements or vision based trajectory planning. This allows us to create annotations (positive and negative samples) for many different tasks, which are exhaustive and expensive in reality.

\begin{figure}[h!]
    \centering
\begin{minipage}{0.25\textwidth}
  \centering
\includegraphics[width=1.0\textwidth]{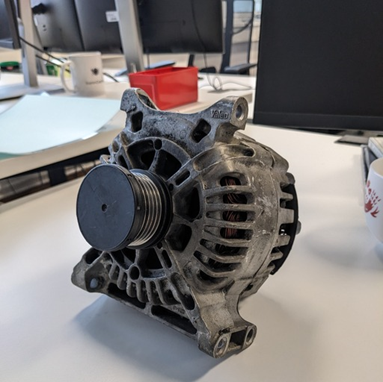}
\subcaption{Input Image: \textbf{Motor}}
\end{minipage}
\begin{minipage}{0.70\textwidth}
    \begin{minipage}{1.0\textwidth}
    \centering
    \includegraphics[width=1.0\textwidth]{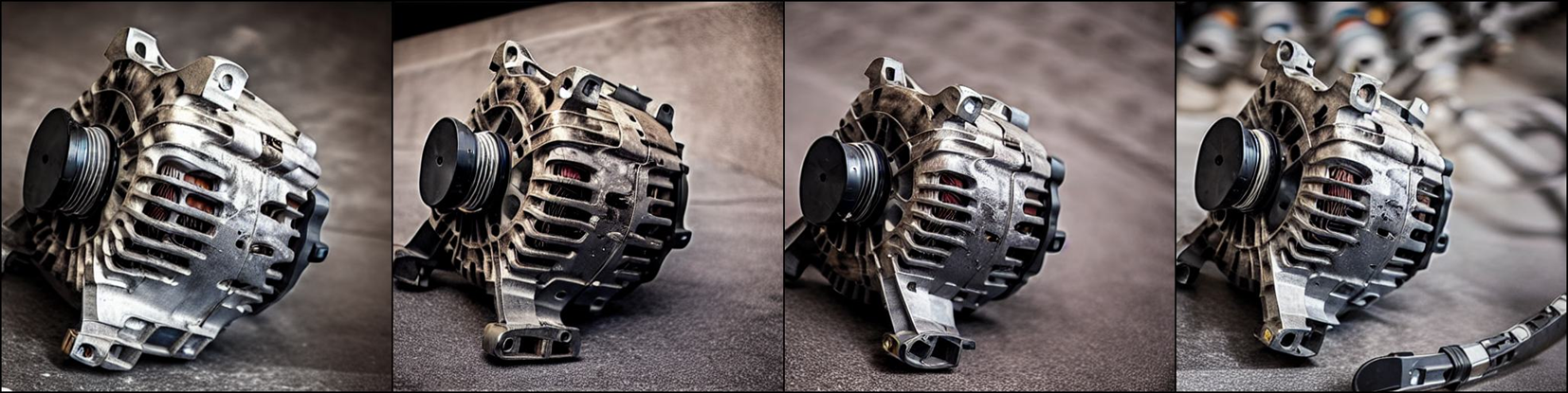}
    \subcaption{Prompt: An Image of a \textbf{motor}.}
    \end{minipage}
    \begin{minipage}{1.0\textwidth}
      \centering
    \includegraphics[width=1.0\textwidth]{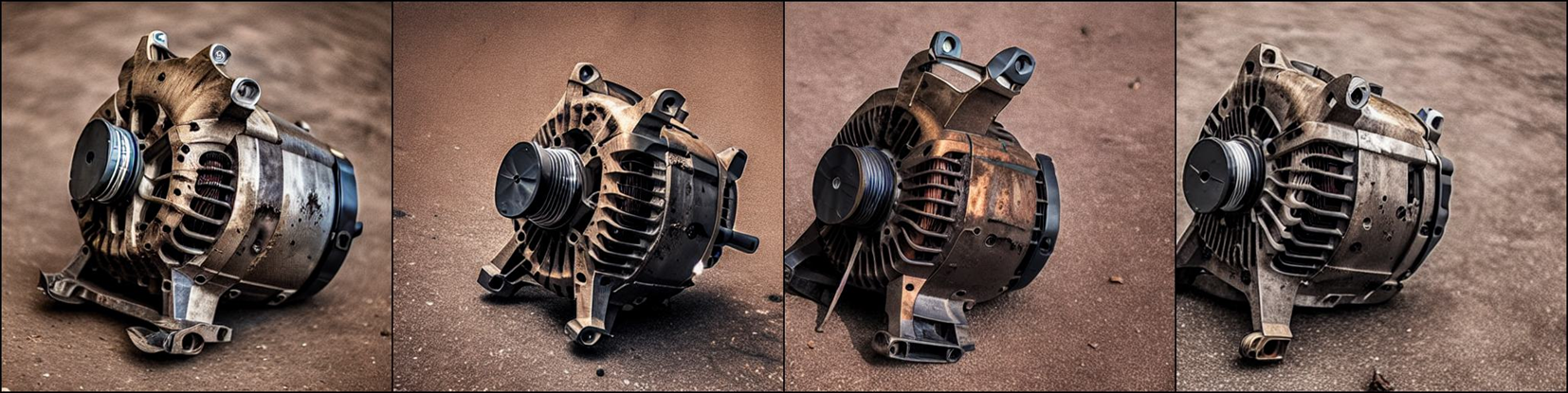}
    \subcaption{Prompt: An Image of a \textbf{motor with a rusty surface}}
    \end{minipage}
\end{minipage}
\caption{Image augmentation with Perfusion~\cite{perfusion}. First experiments for natural language based editing of the visual appearance of assets (before or after simulation).}
\label{fig:perfusion}
\end{figure}

\textbf{(4) Train AI-Helpers:}
Rather then training AI-models to solve a given downstream task from the simulation data alone, it is has been found to be very beneficial to incorporate real data along site of the synthetic data, in order to close the sim2real gap~\cite{SSL6D,kochSyntDataMMM}. Annotating the real world data can be at times very exhausting, especially for robotic related downstream task. Therefore, we propose to train the AI models iterative and step-by-step in order to help to link the simulated data back to the real world, where the 3D assets originated from. In particular we aim e.g. to use 6D pose estimation models trained from simulated data given the 3D assets from real world scenes in order to initialize poses of those real assets in their real world scenes. This in turn allows us to deploy further registration onto the scenery scanning in order to increase the precision of the annotated the real world data. From this linkage we can then e.g. update the training data for the pose estimation and further importantly bring the real world scene into the simulation in order to find possible grasping poses or execute and validate planning strategies. Once the simulated experiments are conducted we have a annotated real world scene with real world data from which we can train novel end2end solutions (grasping pose estimation or planning). Other possible AI-helpers which we can train in this scenery are style transfers~\cite{StyleTransfer} and texture asset augmentations~\ref{fig:perfusion}. Moreover, we aim to combine simulation trained segmentation models with generalized models~\cite{DinoV2,segAny} for identifying overlaps for masking annotations with a linkage to the asset knowledge (e.g. the placement area).




\section{Conclusion}
In this paper we outlined the current trends in vision-based cognitive robotics for object manipulation. Based on those trends we identified the need for a large amount of well annotated training data from real and simulated environments. With our proposed linkage of real world scenes with simulations we aim to provide such data, iterative bridge the sim2real domain gap, and create a training foundation for many AI-tool for  cognitive robotics. 

%
%

\small
\bibliographystyle{ieeetr} 
\bibliography{main}

\end{document}